# Assessing the Performance Gap Between Lexical and Semantic Models for Information Retrieval With Formulaic Legal Language

Larissa Mori
Purdue University
West Lafayette, Indiana, USA

Carlos Sousa de Oliveira
Purdue University
West Lafayette, Indiana, USA

Yuehwern Yih
Purdue University
West Lafayette, Indiana, USA

Mario Ventresca
mventresca@purdue.edu
Purdue University
West Lafayette, Indiana, USA

## Abstract

Legal passage retrieval is an important task that assists legal prac-titioners in the time-intensive process of finding relevant prece-dents to support legal arguments. This study investigates the task of retrieving legal passages or paragraphs from decisions of the Court of Justice of the European Union (CJEU), whose language is highly structured and formulaic, leading to repetitive patterns. Understanding when lexical or semantic models are more effec-tive at handling the repetitive nature of legal language is key to developing retrieval systems that are more accurate, efficient, and transparent for specific legal domains. To this end, we explore when this routinized legal language is better suited for retrieval using methods that rely on lexical and statistical features, such as BM25, or dense retrieval models trained to capture semantic and contex-tual information. A qualitative and quantitative analysis with three complementary metrics shows that both lexical and dense models perform well in scenarios with more repetitive usage of language, whereas BM25 performs better than the dense models in more nuanced scenarios where repetition and verbatim quotes are less prevalent and in longer queries. Our experiments also show that BM25 is a strong baseline, surpassing off-the-shelf dense models in 4 out of 7 performance metrics. However, fine-tuning a dense model on domain-specific data led to improved performance, surpassing BM25 in most metrics, and we analyze the effect of the amount of data used in fine-tuning on the model's performance and temporal robustness. The code, dataset and appendix related to this work are available on: https://github.com/larimo/lexsem-legal-ir.

## 1 Introduction

While information retrieval in the legal domain has long been a focus of sustained research, recent years have seen growing activity, reflected in the release of new datasets, shared tasks, and survey pa-pers aimed at legal case retrieval [4, 8, 12, 23]. In this context, a num-ber of works have focused on the task of legal passage/paragraph retrieval [13, 16, 21]. Instead of focusing on retrieving an entire legal case as a relevant document given a query [4], the goal is to retrieve relevant passages (e.g., text snippets or paragraphs) from past cases given a query, itself also a passage.

Focusing on legal passage retrieval instead of the entire case has the following advantages: (i) a legal case may discuss multiple topics (e.g., procedural aspects, such as how the case was handled; and substantive aspects, such as the actual legal issues at stake), so a recommendation at the case level is not specific enough for a given topic; (ii) legal practitioners can focus on more specific legal concepts before reading the whole document, thus saving time when parsing results [13]; (iii) on the computational side, it avoids the computational difficulties of processing long documents, especially for BERT-based models [3], whose context window may not fit the text of the entire decision [14, 21].

In this work, we focus on the task of legal passage/paragraph retrieval for the Court of Justice of the European Union (CJEU), which is the primary judicial authority of the European Union. One important feature of the CJEU is that the references to a cited legal case point to specific paragraphs in other decisions (see examples of such references in yellow in Figure 1), versus the common practice in other courts (e.g., the US Supreme Court) of citing the entire case. This pattern in the CJEU renders legal passage/paragraph retrieval a natural task, where the goal is to, given a paragraph as a query, retrieve the paragraphs cited in the query from a historical database of previous paragraphs [16]. Notably, in CJEU citations the citing paragraph tends to be a verbatim quote or close paraphrase of the cited paragraph (see example of a citation pair in Table 1).

---

[1]A preliminary version of the work was accepted as a non-archival submission and presented in the 6th Natural Legal Language Processing (NLLP) Workshop co-located with Empirical Methods in Natural Language Processing (EMNLP) 2024.

| 53 | As a preliminary point, it should be borne in mind that the objective pursued by the GDPR, as is set out in Article 1 thereof and in recitals 1 and 10 thereof, consists, inter alia, in ensuring a high level of protection of the fundamental rights and freedoms of natural persons, in particular their right to privacy with respect to the processing of personal data, as enshrined in Article 8(1) of the Charter and Article 16(1) TFEU (judgment of 4 May 2023, *Bundesrepublik Deutschland* (Court electronic mailbox), C-60/22, EU:C:2023:373, paragraph 64). |
| 64 | In the third and last place, the literal interpretation of the GDPR set out in paragraph 61 of the present judgment is supported by the objective pursued by that regulation, as set out in Article 1 thereof and recitals 1 and 10 thereof, which consists, inter alia, in ensuring a high level of protection of the fundamental rights and freedoms of natural persons, in particular their right to privacy with respect to the processing of personal data, as enshrined in Article 8(1) of the Charter of Fundamental Rights of the European Union and Article 16(1) TFEU. (see, to that effect, judgment of 1 August 2022, *Vyriausioji tarnybinės etikos komisija*, C-184/20, EU:C:2022:601, paragraph 125 and the case-law cited). |

**Figure 1: Example of a citing paragraph (top) and a cited paragraph (bottom) in CJEU decisions. Each paragraph is numbered and each reference to a cited case points to a specific paragraph (in yellow). In bold, we highlight the high lexical overlap between the citing and cited paragraphs. For the passage retrieval task, given the citing paragraph, the goal is to retrieve the cited paragraph from a historical database of previous paragraphs.**

Another relevant characteristic of the citations in the CJEU is the adoption of *judicial formulas* when citing previous cases [1, 17]. These are legal phrases[1] that the CJEU repeats as self-standing statements of the law or uses with other prefabricated phrases to ensure uniformity between citations, speed up the writing process, facilitate search of similar cases, etc. [17]. This, along with the aforementioned pattern of verbatim citations, makes the language of the CJEU especially routinized, as compared to the language of courts in France, Germany and the United Kingdom [22].

Given these characteristics of CJEU decisions, we investigate when such formulaic citation structure favors retrieval by simple statistical methods (based on bag of words, term frequency, etc.), that capture such lexical regularities. More specifically, we analyze how strong BM25 is as a baseline for the task of passage retrieval when compared to dense retrieval models. We choose the BM25 method as it has shown to be a strong baseline in other legal case retrieval benchmark datasets from other jurisdictions [20] and is recognized in the information retrieval literature for its strong performance [10, 11, 19]. A qualitative and quantitative analysis using three complementary metrics reveals that, when BM25 outperforms dense models, there is less overlap between the query and the target passage, with statistical significance (Section 5). Thus, BM25 seems better able to capture the relevance from the vocabulary overlap, while dense models may miss the semantic similarity due to bias or noise from additional information in the context beyond the direct quote.

Determining when statistical models are more effective than dense models at managing the repetitive patterns of legal language can guide the creation of more precise and efficient retrieval systems designed for particular legal domains. Given that BM25 is an unsupervised method based on term frequency, while dense retrieval models are trained on large datasets with a high number of parameters and significant computational costs, exploring their trade-offs is critical for optimizing retrieval tasks in resource-constrained environments.

## 2 Related work

### 2.1 Legal passage retrieval

Recent years have seen a rising interest in the task of legal passage retrieval, due to the continuous increase in the number of decisions and the need for legal practitioners to more efficiently perform legal research [13, 21].

The existing datasets for the task of legal passage retrieval vary in terms of the format of the query and the predicted output. LePaRD [13] is a dataset for the task of legal passage retrieval for U.S. federal judgements, where the the goal is to find passages in previous cases given a new passage. CLERC [8], in turn, is another dataset of U.S. federal judgments, set up for the goal of retrieving the appropriate case citations given a passage, instead of retrieving a passage as output. A recent dataset for legal case retrieval was also released for the European Court of Human Rights [23], but the task is to retrieve entire cases, given the factual description of a case.

To the best of our knowledge, the dataset provided in [16] for the CJEU is the first to address the problem of legal passage/paragraph retrieval in European Union Law. Given its recent release, the baselines are limited to the variants of 3 methods and a single evaluation metric in the original paper. Given the relevance of the task and the interest of assessing BM25 as a method, here we extend the set of baselines and evaluation metrics (detailed in Section 3.2), and analyze the implications of the repetitive language of the CJEU's citations on the task of legal passage retrieval for lexical and semantic models.

## 3 Experimental setup

### 3.1 Dataset

The dataset used in our experiments consists of legal judgments from the Court of Justice of the European Union (CJEU) and was published in 2023 in [16]. It is composed of paragraph-to-paragraph citations, manually extracted[2] from legal cases decided by the court between February 1978 and October 2021. We further pre-processed the original dataset to remove 269 paragraphs that were missing text, and 4,851 paragraphs in French to avoid biases in the evaluation metrics due to language differences. We provide the statistics in Table 1.

We split the paragraphs in the dataset into training, validation and test sets, according to a temporal criterion. The periods were chosen so that the number of citations in training, validation and test were approximately 80%, 10%, and 10% of the total, respectively. For each paragraph in the test set, the set of candidate nodes are all of the paragraphs in the training and validation sets[3]. The statistics of the dataset splits can be found in Table 2. We release the preprocessed dataset and the dataset splits in: <suppressed for review>.

---

[1]Example of a judicial formula used by the CJEU about the applicability of EU law: "Whilst it is not in dispute that EU law does not detract from the powers of the Member States (...), the fact remains that, when exercising those powers, the Member States must comply with EU law" [1].

[2]The CJEU cases were obtained from EUR-LEX, which is the official online database of EU Law, and the the paragraphs and citations were extracted from XML/HTML markup [16]. Dataset available under Apache 2.0 license.

| | |
|---|---|
| Unique decisions | 9,651 |
| Unique paragraphs | 83,503 |
| Average of paragraphs per decision | 8.48 (7.63) |
| Average words per paragraph | 106.49 (76.93) |
| Paragraph citations | 102,507 |
| Average inbound citations | 2.1 (2.35) |
| Average outbound citations | 1.88 (1.85) |

**Table 1: Dataset statistics. For averages, the standard deviation is shown in parenthesis. The number of average inbound and outbound citations are shown for paragraphs that receive at least one citation and cite at least one paragraph, respectively.**

The ground truth regarding which paragraphs are relevant given a query paragraph is taken to be those paragraphs that are cited by the query paragraph. A limitation of the dataset is that only paragraphs that cite or are cited appear in the dataset, resulting in the absence of information regarding the further context around the citing and cited texts, which could be useful for the development or assessment of retrieval methods. As a result, the reported results in Section 4 represent a lower bound on performance, since incorporating additional contextual information around the citing and cited paragraphs might enable more effective retrieval.

| Split | Period | # citations | # paragraphs |
|---|---|---|---|
| Training | 1979-2016 | 83,953 | 67,842 |
| Validation | 2017-2018 | 11,883 | 8,973 |
| Test | 2019-2021 | 10,396 | 5,052 |

**Table 2: Statistics for training, validation and test splits.**

## 3.2 Methods

### 3.2.1 Lexical models. **BM25** The BM25 score between a query $q$ and a document $d$ is calculated by summing the contributions of each query term that occurs in the document:

$$\text{BM25}(q, d) = \sum_{t \in q \cap d} \log\left(\frac{N - \text{df}(t) + 0.5}{\text{df}(t) + 0.5}\right) \cdot$$
$$\frac{\text{tf}(t, d) \cdot (k_1 + 1)}{\text{tf}(t, d) + k_1 \cdot \left(1 - b + b \cdot \frac{l_d}{L}\right)} \quad (1)$$

where the log term represents the inverse document frequency (IDF), where $N$ is the total number of documents in the corpus, and $df(t)$ indicates the document frequency of term $t$. The second term of the summation measures how frequently a term appears in the document and adjusts for the length of the document[4], where

$tf(t, d)$ denotes the term frequency of term $t$ in document $d$, $l_d$ represents the length of document $d$, and $L$ is the average document length of all documents in the collection.

**TF-IDF** We include TF-IDF as another bag-of-words baseline that combines term frequency and inverse document frequency to compute the relevance of terms. In comparison to BM25, TF-IDF does not account for the term frequency scaling or document length normalization. We represent the documents with the top-K frequent N-grams of the training set, where $K = 5,000$ and $N \in [1, 2]$.

### 3.2.2 Dense models. **Zero-shot dense models** We compare five zero-shot (i.e., not fine-tuned on the CJEU dataset) dense retrieval methods. The first two models[5] are SBERT [18] and SimCSE [5], which were the only neural baselines considered in the original paper. Both are BERT-based models [3] that were fine-tuned with contrastive learning objectives for sentence representation. The other three models are Nomic's `nomic-embed-text-v1.5` [15] and OpenAI's `text-embedding-ada-002` (Ada-v2) and `text-embedding-3-large` (Emb-3-large)[6], which are models specialized for text embeddings.

**Fine-tuned dense models** We also consider two versions of the SBERT [18] model fine-tuned[7] on the CJEU dataset, SBERT-ft and LegalSBERT-ft. For the SBERT-ft model we fine-tuned the `all-MiniLM-L6-v2` (33M parameters) model with contrastive learning over the training dataset. For LegalSBERT-ft, we similarly fine-tuned LegalBERT[8] [2], a 110M parameter BERT-based model pretrained on English language legal texts.

For each of the models, given a query, we compute the cosine similarity between the embeddings of the query and that of each candidate citation as that candidate's relevance score (or the pairwise score, in the case of BM25). We leave for future work the comparison of efficient retrieval methods such as FAISS [9].

We thus extend the number of models being compared from 6 to 10, as contrasted to the paper that released the dataset [16]. Our analysis encompasses 2 variants of lexical models (BM25 and TF-IDF), as opposed to their consideration of only unigram TF-IDF, and we consider variants of 6 dense models (SBERT, SimCSE, Nomic, Ada-v2, and Emb-3-large), as opposed to their consideration of variants of only 2 dense models, SBERT and SimCSE. Table 8 summarizes the number of parameters and the embedding dimension in each of the dense models.

## 3.3 Evaluation metrics

For the evaluation metrics, we use Recall@k, for $k = 1, 5, 10, 20$, nDCG@10, the Mean Average Precision (MAP), and the Mean Reciprocal Rank (MRR), which are standard metrics used in other legal case retrieval datasets [8, 13]. For all of the metrics, the interpretation is that the higher the value the better the performance of the evaluated model. The Recall@k metric measures the proportion of target paragraphs (i.e., those cited in the query paragraph) that are ranked among the top k retrieved passages by a given method. For instance, under Recall@20, if the query paragraph contains two citations and only one is retrieved within the top 20 passages, the score would be 0.5. The recall metrics are important to estimate the quality of the top k results that would be displayed to a lawyer on a search page, the nDCG@10 metric provides information about the ranking of the top 10 results, and the MAP and MRR metric provides a metric associated to the global ranking of the ground truth documents. This extends the number of metrics taken into consideration in the paper that released the dataset [16] from 1 to 7, which only took into account MAP.

---

[3]We removed the citations between paragraphs in the test set, so that there are no dependencies among paragraphs in the test set.

[4]We note that $k_1$ and $b$ are free parameters, where $k_1$ controls the term frequency scaling and $b$ adjusts the influence of document length on term frequency. In this work, we use the default values $k_1 = 1.2$ and $b = 0.75$.

[5]We used the following models for SBERT [18] and SimCSE [5], respectively: `all-MiniLM-L6-v2` and `sup-simcse-roberta-base` from HuggingFace. For SBERT, we also tested `mpnet-base-v2`, but `all-MiniLM-L6-v2` had better performance for all evaluation metrics considered.

[6]We note that both the Ada-v2 and Emb-3-large models are proprietary, requiring text to be sent via an API for embedding generation, which restricts their safe usage for sensitive information. However, we include these models as references of proprietary, state-of-the-art text embedding models.

[7]We employ the Multiple Negatives Ranking Loss [7] and the default hyperparameter settings of the Sentence Transformers library and codebase to fine-tune both models.

[8]Available at https://huggingface.co/nlpaueb/legal-bert-base-uncased.

| Method | # params. | embed. dim. |
|---|---|---|
| SBERT/SBERT-ft | 33M | 384 |
| LegalSBERT-ft | 110M | 768 |
| SimCSE | 125M | 768 |
| Nomic | 137M | 768 |
| Ada-v2 | N/A | 1,536 |
| Emb-3-large | N/A | 3,072 |

**Table 3: Number of parameters and embedding dimension in each of the dense models employed in our analysis. N/A stands for Not Available, since Ada-v2's and Emb-3-large's counts of parameters are not made public by OpenAI.**

## 4 Performance comparison of lexical and dense models

We display the results for the performance comparison of the lexical and dense models in Table 4.

**Lexical models** BM25 has better performance than TF-IDF's 1-gram and 2-gram versions for all metrics. For example, for Recall@5, BM25 has a performance of 63.63%, compared to 61% for TF-IDF-1gram. Nonetheless, TF-IDF is still a strong baseline compared to the zero-shot dense models, where it outperforms (in both versions) all such models for all metrics, except for Ada-v2. Ada-v2 outperforms TF-IDF in all metrics, except in Recall@10 and Recall@20. For example, TF-IDF-1gram has a Recall@20 of 76.74% and Ada-v2 of 75.47%.

**BM25 outperforms most zero-shot dense models** The experimental results show that BM25 is a strong baseline in comparison with the zero-shot dense models for this task. It outperforms all such models in Recall@K, for K = 5, 10, 20, and the nDCG@10 metric, only being surpassed by Ada-v2 in Recall@1, MAP and MRR. For Recall@5, for instance, BM25 achieves a performance of 63.63%, whereas the best zero-shot model was the Ada-v2 model with 61.28%. On the other hand, Ada-v2 outperforms BM25 with a Recall@1 of 37.3% versus 35.47% for BM25, for example.

Among the zero-shot dense models, the Ada-v2 model outperforms all the others, including one of its successors Emb-3-large, for all metrics in our assessment. We note that the ranking of the zero-shot dense models' performance is consistent across all the metrics analyzed in our experiment. That is, a zero-shot model that is better than another model in one metric is also better than all such models in the other metrics considered.

**Fine-tuned dense models** Across all metrics, the larger LegalSBERT-ft model outperforms BM25 and all other methods in our assessment, with the smaller SBERT-ft model displaying a strong performance and achieving the second best results for all but Recall@20 (where SBERT-ft achieves 79.42% versus 80.43% for BM25).

It is worth noting that SBERT-ft has 33 million parameters and outperforms models four times its size (e.g., Nomic with 137 million parameters). This highlights the importance of fine-tuning dense models on this specific task and dataset for improved performance. We note, however, that fine-tuning is resource-intensive in terms of computational requirements and energy, and does not necessarily yield improvements over much simpler baselines such as BM25, depending on the metrics considered, such as Recall@20 with SBERT-ft in this experiment.

## 5 Analyzing performance gaps between BM25 and dense models

### 5.1 Qualitative analysis

We examine the characteristics of the examples in which BM25 (a method based primarily on term frequency) outperforms dense models that have millions of trainable parameters meant to better capture contextual information on large amounts of textual data. We use Ada-v2 as the representative method for the zero-shot dense models, as it was the best performing among them for all metrics in our experiments, and the SBERT-ft as the fine-tuned dense model whose performance surpasses but is closer to that of BM25.

**BM25 vs. Ada-v2** We randomly sample 3 (out of 359) citation-pair documents where BM25 has 100% recall and Ada-v2 has zero recall. The purpose is to examine the characteristics of the examples in which BM25 performs well and Ada-v2 performs poorly. For comparison, we also sample 3 (out of 2,292) citation-pair documents in which both methods have 100% recall. We use Recall@5 as the recall metric. An example of each scenario is shown in Table 5 and the documents retrieved are shown in the Appendix. In both cases, the direct quotes and/or paraphrases are highlighted in pink for visualization of the portion of the texts that overlap in the query and the cited paragraphs.

For the citation-pair documents in which both methods achieve 100% recall, we observe that the text in the query document is, in its entirety, either a direct quote or a close paraphrase of (parts of the text of) the cited document. For instance, in an example of perfect recall for both methods, the text of paragraph 22 of Caviro Distillerie Srl and Others v European Commission is exactly the same as the cited paragraph 55 of Bricmate AB v Tullverket (Example 1 in Table 9). The other examples retrieved follow the same pattern, where the query is a paraphrase of the cited document. For instance, in Example 3 in Table 9, both documents differ only on small changes, such as stating "The concept of 'seller or supplier'" versus "It follows that the notion of 'seller or supplier'", while the remainder of the text is the same.

In the citation-pair documents retrieved when BM25 has 100% recall and Ada v2 has zero recall, the same pattern is observed to a lesser degree. There is no query document that is a direct quote or a close paraphrase of the cited document in its entirety. In the retrieved Examples 1-3 in Table 10, the queries contain a paraphrase of parts of the text of the cited paragraph, but the quote is at least less than half of the content of the queries. For instance, in Example 2 in Table 10, the query spans multiple lines of text, but only the last two lines are a direct quote of the cited paragraph.

| Method | R@1 | R@5 | R@10 | R@20 | nDCG@10 | MAP | MRR |
|---|---|---|---|---|---|---|---|
| BM25 | 0.3547 | *0.6363* | *0.7289* | <u>0.8043</u> | *0.5767* | 0.5132 | 0.5786 |
| TF-IDF-1gr | 0.3479 | 0.6108 | 0.6992 | 0.7674 | 0.5579 | 0.4847 | 0.5640 |
| TF-IDF-2gr | 0.3416 | 0.6004 | 0.6885 | 0.7586 | 0.5491 | 0.4829 | 0.5555 |
| SBERT | 0.3136 | 0.5210 | 0.5883 | 0.6486 | 0.4797 | 0.4325 | 0.4921 |
| SimCSE | 0.2745 | 0.4455 | 0.4996 | 0.5502 | 0.4065 | 0.3647 | 0.4223 |
| Nomic | 0.3142 | 0.5406 | 0.6132 | 0.6762 | 0.4930 | 0.4411 | 0.5025 |
| Ada-v2 | *0.3730* | 0.6128 | 0.6911 | 0.7547 | 0.5696 | *0.5150* | *0.5814* |
| Emb-3-large | 0.3358 | 0.5869 | 0.6726 | 0.7431 | 0.5357 | 0.4778 | 0.5431 |
| SBERT-ft | <u>0.4180</u> | <u>0.6729</u> | <u>0.7428</u> | <u>0.7942</u> | <u>0.6266</u> | <u>0.5724</u> | <u>0.6391</u> |
| LegalSBERT-ft | **0.4202** | **0.6859** | **0.7607** | **0.8179** | **0.6376** | **0.5814** | **0.6486** |

**Table 4: Baseline metrics for the CJEU test set for the lexical models and dense models. TF-IDF-$ngr$ indicates $n$ as the number of successive words considered. The metrics considered are Recall@K (R@K), $k = 1, 5, 10, 20$, nDCG@10, MAP and MRR. We highlight in bold the best performing model, <u>underline the second best</u>, and *italicize the third best* for each metric.**

| | Query document | Cited paragraph | Metrics |
|---|---|---|---|
| **Both perfect recall** | Finally, with respect to the causal link, under Article 3(6) of the basic regulation the EU institutions must demonstrate that the volume and/or price levels identified pursuant to Article 3(3) are responsible for an impact on the Union industry as provided for in Article 3(5) and that that impact exists to a degree which enables it to be classified as material (judgment of 10 September 2015, Bricmate, C-569/13, EU:C:2015:572, paragraph 55). | Finally, with respect to the causal link, under Article 3(6) of the basic regulation, the EU institutions must demonstrate that the volume and/or price levels identified pursuant to Article 3(3) are responsible for an impact on the Union industry as provided for in Article 3(5) and that that impact exists to a degree which enables it to be classified as material. | Mean edit distance: 18 3-grams in common: 68 4-grams in common: 69 LCS: 73 |
| **BM25 > Ada v2** | In point 1 of the operative part of that judgment, the Court ruled that Article 325(1) TFEU precludes national legislation that establishes a procedure for the termination of criminal proceedings (...). It added, in the same point, that it is for the national court to give full effect to Article 325(1) TFEU, by disapplying that legislation, where necessary, while also ensuring respect for the fundamental rights of the persons accused, stating, in paragraph 70 of that judgment, that those rights include the right of those persons to have their case heard within a reasonable time. | 70 In the second place, the referring court must, when it decides on the measures to be applied in this specific case in order to give full effect to Article 325(1) TFEU, protect the right of accused persons to have their case heard within a reasonable time. | Mean edit distance: 129 3-grams in common: 20 4-grams in common: 17 LCS: 35 |

**Table 5: Examples of pairs of query and cited document when both BM25 and Ada-v2 achieve perfect Recall@5 (top row), and when BM25 has 100% recall and Ada-v2 has 0% Recall@5 (bottom row). In pink, we highlight the parts of the text that are verbatim quotes of the cited paragraph. For each example, we compute the mean edit distance, the number of 3- and 4-gram in common, and the Longest Common Subsequence (LCS).**

**BM25 vs. SBERT-ft** Similar conclusions are drawn when performing the same analysis as above for the fine-tuned dense model SBERT-ft. We randomly sample 3 (out of 103) citation-pair documents where BM25 has 100% recall and SBERT-ft has zero recall, and 3 (out of 2,628) citation-pair documents where both methods have 100% recall. For the recall metric, we use Recall@20, since BM25 outperformed SBERT-ft in this metric.

When both methods have 100% recall, two out of the three examples are verbatim quotes of the cited paragraph in their entirety (Examples 1 and 2 of Table 11), while the third also contains a verbatim quote of the cited paragraph (Example 3 of Table 11).

In contrast, when BM25 has better performance than SBERT-ft, in two out of three examples the query document does not contain a verbatim quote of (parts of) the cited paragraph. For instance, in Example 1 in Table 12, the query and the cited paragraph only have five short expressions in common (e.g., "the Court has consistently held"), but no substantial overlap as seen in the case of perfect recall for both methods. Example 3 in Table 12 does not contain any overlap of short expressions between citing and cited paragraphs.

While not rigorous, this qualitative analysis highlights that both BM25 and the dense models (Ada-v2 and SBERT-ft) are able to perform well when the text of the query is a verbatim quote (or a close paraphrase) of the text in the cited document. However, when the text of the query is not a verbatim quote but still contains references to the cited text, the BM25 model seems better able to capture the relevance from the vocabulary overlap, while Ada-v2 may miss the semantic similarity due to bias or noise from additional information in the context beyond the direct quote.

To assess the last point that additional information may add noise and decrease Ada-v2's performance, we analyzed the top 5 ranked candidate paragraphs by Ada-v2 for the query in Example 4 in Table 10. The query discusses two topics: (i) Article 4 of the Charter of Fundamental Rights and Article 3 of the European Convention on Human Rights (ECHR); and (ii) Framework Decision 2002/584. While the citation refers to the second topic, Ada-v2 wrongly returned 4 out of 5 paragraphs discussing the first topic, regarding the need for a judicial authority to properly assess the risk of inhuman and degrading treatment of a person subject to an European arrest warrant in Article 4 of the Charter of Fundamental Rights. We provide the top 5 candidate paragraphs in Section D of the Appendix.

## 5.2 Quantitative analysis

We quantitatively verify the pattern observed in our qualitative analysis in Section 5.1, suggestive that the similarity between the query and cited passage correlates with BM25's performance. For this purpose, we use the following complementary metrics as a proxy for the degree to which a query document is a verbatim quote or paraphrase of the cited passage: mean edit distance (at the word level), number of $N$-grams in common, and Longest Common Subsequence (LCS). While the mean edit distance computes the similarity based on the entire text of the passages being compared, $N$-grams focus on the contiguous overlap of terms, and LCS on the non-contiguous overlap, which makes it well-suited to detect paraphrases.

More specifically, the mean edit distance (at the word level) measures the similarity between two sentences by calculating the minimum number of word operations (e.g., replace, delete, etc.) needed to transform one sequence into the other. The number of $N$-grams in common measures the similarity at the level of contiguous sequences of words. The LCS between the query and cited passage refers to the longest sequence of words that appears in both texts in the same order, but not necessarily consecutively.

For all metrics, we performed two-sided t-tests with the null hypothesis that the sample means of the two samples were the same in the scenarios in which BM25 has 100% recall and the dense model has zero recall and in the scenario where both methods achieve perfect recall, at a 5% significance level.

For the sake of brevity, we focus on the performance comparison of BM25 and Ada-v2, and leave the results for the BM25 and SBERT-ft comparison to Section E in the Appendix, since the results and conclusions are aligned in both cases.

**Number of $N$-grams in common** The average number of $N$-grams in common between the query and cited paragraph when BM25 and Ada-v2 have perfect recall is statistically significant higher than when BM25 has 100% recall and Ada-v2 has 0% recall, for all $N = 2, 3, ..., 10$. For example, for $N = 3$, the average number of $N$-grams in common is 56.19 (std. 27.21) in the first scenario and 41.90 (std. 27.75) in the second. For $N = 10$, we observe 35.85 (std. 28.90) $N$-grams in common in the first case compared to 23.14 (std. 27.04) in the second. This indicates a lower degree of verbatim quotes between the query and cited documents in the scenario where BM25 has 100% recall and Ada-v2 has 0% recall. The results are shown in Figure 2.

**Mean edit distance** Similarly, the mean edit distance is statistically significantly higher in the scenarios in which BM25 has 100% recall and Ada-v2 has zero recall versus in the scenario where both methods achieve perfect recall. An average of 98.52 (std. 45.66) word edits are necessary to turn the query into the cited document in the first scenario compared to 66.81 (std. 39.84) in the second. The results are shown in Figure 3.

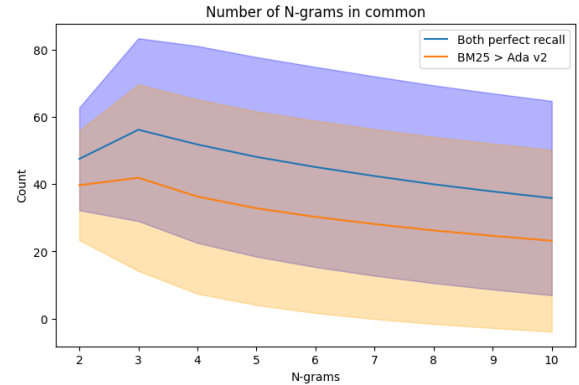

**Figure 2: Average number of $N$-grams in common between query and cited paragraph when both BM25 and Ada-v2 have perfect Recall@5 (*Both perfect recall*), and when BM25 has 100% recall and Ada-v2 has 0% Recall@5 (*BM25 > Ada-v2*), for $N = 2, 3, ..., 10$. The shaded regions indicate the standard deviation.**

**LCS** Consistently with the other metrics, the average LCS is also ~14% higher in the scenario of perfect recall, where the comparison is also statistically significant. This implies that, in the scenario where BM25 outperforms Ada-v2, the query and cited documents have a lower overlap of non-contiguous words, a proxy for the number of paraphrases. When both methods have perfect recall, the average LCS is 67.59 (std. 30.97), while the average is 53.68 (std. 30.8) when BM25 has better performance. The results are shown in Figure 3.

The three metrics analyzed point to the same conclusion that there is a lower presence of verbatim quotes and paraphrases between the query and cited documents in the scenario in which BM25 performs well and Ada-v2 does not. Thus, both lexical and dense models perform well in the scenarios with more repetitive usage of language, whereas BM25 performs better than the dense models in more nuanced scenarios where repetition and verbatim quotes are less prevalent.

There is also evidence that BM25 has better performance for longer queries, where dense models may miss the quote relevance due to noise resultant from the additional information. When BM25 has better performance, the average number of words in each query is 102.37 (std. 37.34), and the median is 96 words. When both models have perfect recall, the average number of words is 94.33 (std. 34.24), with a median of 88 words. The difference between the mean of the samples is statistically significant at 5%.

## 6 Performance comparison for different amounts of data in fine-tuning

We perform an ablation study to evaluate the effect of the amount of data used in fine-tuning on the models' performance and also their temporal robustness (such as their ability to generalize to distant time periods) in a corpus with repetitive and formulaic language. Specifically, we fine-tune the SBERT-ft model on different percentages $p$% obtained from a chronological ordering of the data, where $p \in \{10, 20, 30, 50, 70, 80\}$. For each model trained on the different fine-tuning percentages, we compute the evaluation metrics on the test set (2019-2021 period). The results, including the years in training and validation used for fine-tuning each SBERT-ft model and their performance on the recall metrics, are shown in Table 6. Due to space constraints, we leave to Section F in the Appendix the results for nDCG@10, MAP and MRR, since the conclusions are aligned in both cases.

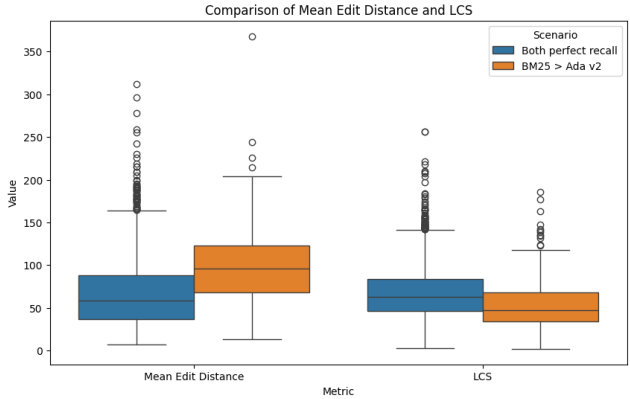

Figure 3: Comparison between Mean Edit Distance and LCS between query and cited paragraph when both BM25 and Ada-v2 have perfect Recall@5 (*Both perfect recall*), and when BM25 has 100% recall and Ada-v2 has 0% Recall@5 (*BM25 > Ada-v2*). On average, the mean edit distance is lower and the LCS is higher when both methods have perfect recall, indicating a higher degree of similarity between the documents than when BM25 outperforms Ada-v2. Both results are statistically significant.

| Training | Validation | R@1 | R@5 | R@10 | R@20 |
|----------|-----------|-----|-----|------|------|
| Zero-shot | - | 0.3136 | 0.5210 | 0.5883 | 0.6486 |
| **10%** 1978-2000 | 2001-2004 | 0.3980 | 0.6421 | 0.7127 | 0.7676 |
| **20%** 1978-2004 | 2005-2007 | 0.4141 | 0.6619 | 0.7266 | 0.7816 |
| **30%** 1978-2007 | 2008-2010 | 0.4184 | 0.6613 | 0.7314 | 0.7823 |
| **50%** 1978-2011 | 2012-2014 | **0.4213** | **0.6735** | <u>0.7412</u> | 0.7917 |
| **70%** 1978-2015 | 2016-2017 | <u>0.4212</u> | 0.6708 | 0.7391 | 0.7917 |
| **80%** 1978-2016 | 2017-2018 | 0.4180 | <u>0.6729</u> | **0.7428** | <u>0.7942</u> |
| BM25 | - | 0.3547 | 0.6363 | 0.7289 | **0.8043** |

Table 6: Evaluation metrics for different percentages of data used for fine-tuning the SBERT-ft model, with years used for training and validation. The test set is the split for 2019-2021 years. We highlight in bold and <u>underline</u> the best and second-best models, respectively, for each metric. The results show that the models fine-tuned on more data tend to perform better on the test set, but performance increases tend to plateau as the amount of data increases.

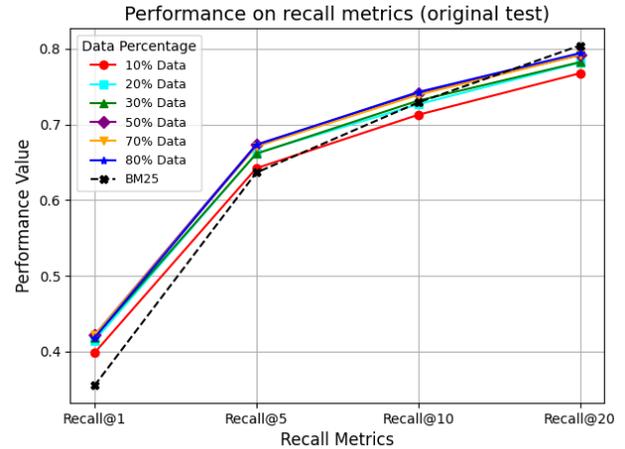

Figure 4: Performance on Recall@k metrics, $k = 1, 5, 10, 20$, for SBERT-ft fine-tuned with different data percentages and BM25 baseline on the test set (2019-2021).

**Increasing training data generally improves performance in the original test set** For all metrics, as the amount of training data increases, the performance of the models tends to increase. For example, for Recall@5, the performance increases from 64.21% to 67.29% from fine-tuning on 10% of data to 80% of data, which is a 4.8% relative (3.08% absolute) increase. However, the performance increase is larger from zero-shot to fine-tuning than from increasing the amounts of data. From zero-shot to 10% fine-tuning, the relative (absolute) increase is 16.9% (8.8%) for Recall@5. This highlights that, in scenarios of highly repetitive and formulaic language, fine-tuning is still helpful in increasing performance even when there is less data available.

**Higher percentages do not necessarily outperform lower percentages.** For instance, the Recall@1 from fine-tuning on 50% of the data is 42.13%, which is higher than the performance of 41.8% for 80% of data. It is worth noting that the differences in performances may be affected by randomness in model training (e.g., batch samples) and not relevant in a real-world scenario. However, this is evidence that performance increases plateau after using a certain amount of data in this setting.

**Comparison with BM25** The fine-tuned SBERT-ft models outperform the BM25 baseline at most data percentages, especially at higher fine-tuning percentages like 70% and 80%. As the recall levels increase, the ranking of BM25 relative to the fine-tuned models improves. More specifically, BM25 has the lowest performance compared to all different fine-tuning percentages for Recall@1 but approximately matches the performance of the 30% fine-tuning at Recall@10 and surpasses 80% fine-tuning at Recall@20. BM25 has better performance on coarse-level metrics (e.g., higher Recall@k), while fine-tuned models are more effective for fine-grained metrics (e.g., lower Recall@k). However, BM25 can match or surpass fine-tuned models in recall at lower fine-tuning percentages, indicating that fine-tuning is not always advantageous and depends on data availability. The comparison is shown in Figure 4.

## 7 Conclusion

This study delved into the task of legal passage retrieval within the landscape of highly structured and formulaic language, focusing on the Court of Justice of the European Union (CJEU). Our main objective was to discern when the repetitive nature of CJEU decisions favors lexical methods such as BM25 versus semantic-oriented dense retrieval models.

Our qualitative and quantitative analyses identified that the degree that a query is a verbatim quote of a cited passage as one of the factors that characterize performance gaps between such methods. BM25 serves as a strong baseline outperforming off-the-shelf dense models on several metrics, but its advantage diminishes against dense models fine-tuned on domain-specific data. Our study further indicated how data volume impacts their optimization for performance and temporal robustness, providing insights towards increased efficiency in semantic retrieval in this legal context. Given BM25's demonstrated strength, which echoes its robust baseline performance in other citation-based legal retrieval datasets [20, 23], a valuable next step would be to develop a typology of cases where this lexical approach particularly excels in legal retrieval tasks.

While our study focuses on the CJEU, it raises broader questions about the generalizability of our findings across other judicial bodies. Assessing whether the results for the highly standardized linguistic style of the CJEU generalizes to other courts requires measuring the formulaicity of a court's language, using the frequency of recurring syntactic constructions or the proportion of boiler-plate expressions within a corpus of judicial opinions. In future work, we intend to explore how linguistic and structural variations across legal systems influence retrieval efficacy, thereby situating our findings within a broader legal scholarship context.

Future work could also explore specific methodological advancements. Hybrid approaches that combine BM25's lexical strengths with the semantic capabilities of fine-tuned dense models offer a promising avenue for potentially achieving higher accuracy and robustness. A promising direction for future work is developing a two-stage re-ranking pipeline where BM25 performs initial coarse retrieval, followed by dense model refinement for precision ranking. Further investigation into the specific linguistic features that affect performance in formulaic scenarios could also lead to more targeted improvements in model architecture or training strategies.

## 8 Limitations

In this work, as with much of the legal information retrieval literature, we use citations as a proxy for legal relevance [6, 8, 13, 23]. One of the limitations in practical applications is that the language in citing paragraphs is often influenced by the cited content, including direct quotes or paraphrasing, whereas legal practitioners rarely search using direct quotes or paraphrases from existing texts.

Consequently, using citations as the measure of relevance tends to favor models that are particularly effective at matching lexical patterns between queries and citations. However, such models may not generalize well to practical legal research tasks, where understanding the broader context and nuanced legal concepts is often required. This performance overestimation may be especially pronounced when assessing citations in legal documents produced by courts that adopt very routinized language, such as the CJEU.

Another limitation of the dataset is that only paragraphs that cite or are cited appear in the dataset, resulting in the absence of information regarding the context around the citing and cited texts, which could be useful for the development of methods that take into consideration the broader context of a decision for relevance assessment.

Due to resource constraints, we do not perform a hyperparameter grid search for the SBERT models fine-tuned on CJEU data[9]. Instead, we use the standard hyperparameters in the Sentence Transformers library. Thus, the performance results provide a lower bound on the performance of such models. Although we do not perform a hyperparameter grid search, these models still outperform the other methods considered. In future work, efficient retrieval methods, such as FAISS [9], should also be considered.

## 9 Ethical Considerations

In this study, we compare the performance of lexical and semantic models in information retrieval, utilizing a dataset of historical case law of the CJEU. Both the original dataset and the pre-processed version used in our analysis are publicly available under the Apache 2.0 license, supporting transparency and knowledge sharing. We recognize, however, that judicial decisions reflect the biases and societal norms of their time, which may include content that is now considered objectionable. While our models and experiments are not designed to perpetuate these localized and time-constrained ideologies, the automation of textual analysis carries the risk of inadvertently reinforcing them. Therefore, researchers and practitioners must ensure their work does not contribute to the uncritical reproduction of such biases.

---

[9]For our computations, we use a NVIDIA Geforce RTX 3050 Ti.

## A  Further dataset statistics

We provide further dataset statistics at the decision level. Each decision that has at least one citation has an average of 13.09 outbound citations, with standard deviation of 13.64. Each decision that is cited at least once has an average of 12.31 inbound citations, with standard deviation of 17.6, respectively.

| | |
|---|---|
| Number of unique decisions | 10,456 |
| Average outbound citations per decision | 13.09 (13.64) |
| Average inbound citations per decision | 12.31 (17.6) |

**Table 7: Decision-level statistics. For averages, the standard deviation is shown in parenthesis.**

## B  Model hyperparameters

The number of parameters and the embedding dimension in each of the dense models employed in our analysis is displayed in Table 8. Ada-v2's and Emb-3-large's counts of parameters are not made public by OpenAI. Also, a count of parameters and embedding dimension does not apply to BM25 and TD-IDF, which are thus omitted.

| Method | # params. | embed. dim. |
|---|---|---|
| SBERT/SBERT-ft | 33M | 384 |
| LegalSBERT-ft | 110M | 768 |
| SimCSE | 125M | 768 |
| Nomic | 137M | 768 |
| Ada-v2 | N/A | 1,536 |
| Emb-3-large | N/A | 3,072 |

**Table 8: Number of parameters and embedding dimension in each of the dense models employed in our analysis. Ada-v2's and Emb-3-large's counts of parameters are not made public by OpenAI.**

## C  Qualitative examples of BM25 vs. dense retrieval models

This section provides the examples described in Section 5.1. In both cases, we highlight in pink the direct quotes and/or paraphrases for visualization of the portion of the texts that overlap in the query and the cited paragraphs.

**BM25 vs. Ada-v2** We display the sampled documents for the scenario where (i) BM25 and Ada-v2 have perfect Recall@5 in Figure 9; and (ii) BM25 has 100% recall and Ada-v2 has zero Recall@5 in Table 10, as described in Section 5.1. We highlight in pink the lexical overlap between the citing and cited paragraphs.

**BM25 vs. SBERT-ft** We display the sampled documents for the scenario where (iii) BM25 and SBERT-ft have perfect Recall@20 in Figure 11; and (iv) BM25 has 100% Recall@20 and Ada-v2 has zero Recall@20 in Table 11. We highlight in pink the lexical overlap between the citing and cited paragraphs.

For a side-by-side comparison, we also display an example of each scenario in Table 5, including a comparison of the metrics mean edit distance, number of N-grams in common, and Longest Common Subsequence (LCS).

## D  Qualitative example of top candidates retrieved for Ada-v2

The following are the top 5 candidates selected via Ada-v2 in Example 1 in Table 10:

CELEX: 62015CJ0404
NUMBER: 88
TITLE: Pál Aranyosi and Robert Căldăraru v Generalstaatsanwaltschaft Bremen.
DATE: 2016-04-05

| | Query document | Cited paragraph |
|---|---|---|
| 1 | **CELEX: 62018CJ0345**<br>NUMBER: 22<br>TITLE: Caviro Distillerie Srl and Others v European Commission.<br>DATE: 2019-07-10<br>TEXT: Finally, with respect to the causal link, under Article 3(6) of the basic regulation the EU institutions must demonstrate that the volume and/or price levels identified pursuant to Article 3(3) are responsible for an impact on the Union industry as provided for in Article 3(5) and that that impact exists to a degree which enables it to be classified as material (judgment of 10 September 2015, Bricmate, C-569/13, EU:C:2015:572, paragraph 55). | **CELEX: 62013CJ0569**<br>NUMBER: 55<br>TITLE: Bricmate AB v Tullverket.<br>DATE: 2015-09-10<br>TEXT: 55. Finally, with respect to the causal link, under Article 3(6) of the basic regulation the EU institutions must demonstrate that the volume and/or price levels identified pursuant to Article 3(3) are responsible for an impact on the Union industry as provided for in Article 3(5) and that that impact exists to a degree which enables it to be classified as material. |
| 2 | **CELEX: 62017CJ0590**<br>NUMBER: 36<br>TITLE: Henri Pouvin and Marie Dijoux, v Electricité de France (EDF).<br>DATE: 2019-03-21<br>TEXT: 36 The concept of 'seller or supplier', within the meaning of Article 2(c) of Directive 93/13, is a functional concept, requiring determination of whether the specific contractual relationship is amongst the activities that a person provides in the course of his trade (see, to that effect, judgment of 17 May 2018, Karel de Grote — Hogeschool Katholieke Hogeschool Antwerpen, C-147/16, EU:C:2018:320, paragraph 55). | **CELEX: 62016CJ0147**<br>NUMBER: 55<br>TITLE: Karel de Grote — Hogeschool Katholieke Hogeschool Antwerpen VZW v Susan Romy Jozef Kuijpers.<br>DATE: 2018-05-17<br>TEXT: 55 It follows that the notion of 'seller or supplier', within the meaning of Article 2(c) of Directive 93/13 is a functional concept, requiring determination of whether the contractual relationship is amongst the activities that a person provides in the course of their trade, business or profession (see, by analogy, the order of 27 April 2017, Bachman, C-535/16, not published, EU:C:2017:321, paragraph 36 and the case-law cited). |
| 3 | **CELEX: 62018CJ0706**<br>NUMBER: 29<br>TITLE: X v Belgische Staat.<br>DATE: 2019-11-20<br>TEXT: 29 Under Article 4(1) of Directive 2003/86, Member States are to authorise the entry and residence, in accordance with that directive, of certain members of the sponsor's family for the purpose of family reunification, including, in particular, the sponsor's spouse. The Court has previously held that that provision imposes specific positive obligations, with corresponding clearly defined individual rights, on the Member States, since it requires them, in the cases determined by that directive, to authorise family reunification of certain members of the sponsor's family, without being left a margin of appreciation (judgment of 27 June 2006, Parliament v Council, C-540/03, EU:C:2006:429, paragraph 60). | **CELEX: 62003CJ0540**<br>NUMBER: 60<br>TITLE: European Parliament v Council of the European Union.<br>DATE: 2006-06-27<br>TEXT: 60. Going beyond those provisions, Article 4(1) of the Directive imposes precise positive obligations, with corresponding clearly defined individual rights, on the Member States, since it requires them, in the cases determined by the Directive, to authorise family reunification of certain members of the sponsor's family, without being left a margin of appreciation. |

**Table 9: Examples for scenario (i): BM25 and Ada-v2 with perfect Recall@5. The first column indicates the example number. We highlight in pink the lexical overlap between the citing and cited paragraphs. There is a very high overlap between query document and cited paragraph in this scenario.**

TEXT: 88 It follows that, where the judicial authority of the executing Member State is in possession of evidence of a real risk of inhuman or degrading treatment of individuals detained in the issuing Member State, having regard to the standard of protection of fundamental rights guaranteed by EU law and, in particular, by

Article 4 of the Charter (see, to that effect, judgment in Melloni, C-399/11, EU:C:2013:107, paragraphs 59 and 63, and Opinion 2/13, EU:C:2014:2454, paragraph 192), that judicial authority is bound to assess the existence of that risk when it is called upon to decide on the surrender to the authorities of the issuing Member State of the

| | Query document | Cited paragraph |
|---|---|---|
| 1 | **CELEX: 62018CJ0128**<br>NUMBER: 79<br>TITLE: Dumitru-Tudor Dorobantu.<br>DATE: 2019-10-15<br>TEXT: 79 Last, it should be pointed out that, while it is open to the Member States to make provision in respect of their own prison system for minimum standards in terms of detention conditions that are higher than those resulting from Article 4 of the Charter and Article 3 of the ECHR, as interpreted by the European Court of Human Rights, a Member State may nevertheless, as the executing Member State, make the surrender to the issuing Member State of the person concerned by a European arrest warrant subject only to compliance with the latter requirements, and not with those resulting from its own national law. The opposite solution would, by casting doubt on the uniformity of the standard of protection of fundamental rights as defined by EU law, undermine the principles of mutual trust and recognition which Framework Decision 2002/584 is intended to uphold and would, therefore, compromise the efficacy of that framework decision (see, to that effect, judgment of 26 February 2013, Melloni, C-399/11, EU:C:2013:107, paragraph 63). | **CELEX: 62011CJ0399**<br>NUMBER: 63<br>TITLE: Stefano Melloni v Ministerio Fiscal.<br>DATE: 2013-02-26<br>TEXT: 63. Consequently, allowing a Member State to avail itself of Article 53 of the Charter to make the surrender of a person convicted in absentia conditional upon the conviction being open to review in the issuing Member State, a possibility not provided for under Framework Decision 2009/299, in order to avoid an adverse effect on the right to a fair trial and the rights of the defence guaranteed by the constitution of the executing Member State, by casting doubt on the uniformity of the standard of protection of fundamental rights as defined in that framework decision, would undermine the principles of mutual trust and recognition which that decision purports to uphold and would, therefore, compromise the efficacy of that framework decision. |
| 2 | **CELEX: 62018CJ0704**<br>NUMBER: 17<br>TITLE: Criminal proceedings against Nikolay Boykov Kolev and Others.<br>DATE: 2020-02-12<br>TEXT: 17 In point 1 of the operative part of that judgment, the Court ruled that Article 325(1) TFEU precludes national legislation that establishes a procedure for the termination of criminal proceedings, such as that provided for in Articles 368 and 369 of the Code of Criminal Procedure, in so far as that legislation is applicable in proceedings initiated with respect to cases of serious fraud or other serious illegal activities affecting the financial interests of the European Union in customs matters. It added, in the same point, that it is for the national court to give full effect to Article 325(1) TFEU, by disapplying that legislation, where necessary, while also ensuring respect for the fundamental rights of the persons accused, stating, in paragraph 70 of that judgment, that those rights include the right of those persons to have their case heard within a reasonable time. | **CELEX: 62015CJ0612**<br>NUMBER: 70<br>TITLE: Criminal proceedings against Nikolay Kolev and Others.<br>DATE: 2018-06-05<br>TEXT: 70 In the second place, the referring court must, when it decides on the measures to be applied in this specific case in order to give full effect to Article 325(1) TFEU, protect the right of accused persons to have their case heard within a reasonable time. |
| 3 | **CELEX: 62018CJ0263**<br>NUMBER: 52<br>TITLE: Nederlands Uitgeversverbond and Groep Algemene Uitgevers v Tom Kabinet Internet BV and Others.<br>DATE: 2019-12-19<br>TEXT: 52 In the fourth place, an interpretation of the distribution right referred to in Article 4(1) of Directive 2001/29 as applying only to the distribution of works incorporated in a material medium follows equally from Article 4(2) of that directive, as interpreted by the Court in relation to exhaustion of that right, the Court having ruled that the EU legislature, by using the terms 'tangible article' and 'that object' in recital 28 of that directive, wished to give authors control over the initial marketing in the European Union of each tangible object incorporating their intellectual creation (judgment of 22 January 2015, Art & Allposters International, C-419/13, EU:C:2015:27, paragraph 37). | **CELEX: 62013CJ0419**<br>NUMBER: 37<br>TITLE: Art & Allposters International BV v Stichting Pictoright.<br>DATE: 2015-01-22<br>TEXT: 37. It follows from the foregoing considerations that the EU legislature, by using the terms 'tangible article' and 'that object', wished to give authors control over the initial marketing in the European Union of each tangible object incorporating their intellectual creation. |

**Table 10: Examples for scenario (ii): BM25 has 100% recall and Ada-v2 has zero Recall@5. The first column indicates the example number. We highlight in pink the lexical overlap between the citing and cited paragraphs.**

| | Query document | Cited paragraph |
|---|---|---|
| 1 | **CELEX: 62018CJ0824**<br>NUMBER: 64<br>TITLE: A.B. and Others v Krajowa Rada Sądownictwa and Others.<br>DATE: 2021-03-02<br>TEXT: Second, under the second paragraph of Article 252 TFEU, the Advocate General, acting with complete impartiality and independence, is to make, in open court, reasoned submissions on cases which, in accordance with the Statute of the Court of Justice of the European Union, require the Advocate General's involvement. The Court is not bound either by the Advocate General's submissions or by the reasoning which led to those submissions. Consequently, a party's disagreement with the Opinion of the Advocate General, irrespective of the questions that he or she examines in the Opinion, cannot in itself constitute grounds justifying the reopening of the oral procedure (judgment of 6 March 2018, Achmea, C284/16, EU:C:2018:158, paragraph 27 and the case-law cited). | **CELEX: 62016CJ0284**<br>NUMBER: 27<br>TITLE: Slowakische Republik v Achmea BV.<br>DATE: 2018-03-06<br>TEXT: 27 Secondly, under the second paragraph of Article 252 TFEU, the Advocate General, acting with complete impartiality and independence, is to make, in open court, reasoned submissions on cases which, in accordance with the Statute of the Court of Justice of the European Union, require the Advocate General's involvement. The Court is not bound either by the Advocate General's conclusion or by the reasoning which led to that conclusion. Consequently, a party's disagreement with the Opinion of the Advocate General, irrespective of the questions that he examines in his Opinion, cannot in itself constitute grounds justifying the reopening of the oral procedure (judgment of 25 October 2017, Polbud — Wykonawstwo, C106/16, EU:C:2017:804, paragraph 24 and the case-law cited). |
| 2 | **CELEX: 62018CJ0492**<br>NUMBER: 30<br>TITLE: TC.<br>DATE: 2019-02-12<br>TEXT: 30 In the second place, as regards the criterion relating to urgency, it is necessary, in accordance with the Court's settled case-law, to take into account the fact that the person concerned was deprived of his liberty and that the question as to whether he may continue to be held in custody depends on the outcome of the dispute in the main proceedings. In addition, the situation of the person concerned must be assessed as it stands at the time when consideration is given to the request that the reference be dealt with under the urgent procedure (judgment of 19 September 2018, RO, C327/18 PPU, EU:C:2018:733, paragraph 30 and the case-law cited). | CELEX: 62018CJ0327<br>NUMBER: 30<br>TITLE: Minister for Justice and Equality v RO.<br>DATE: 2018-09-19<br>TEXT: 30 In the second place, as regards the criterion relating to urgency, it is necessary, in accordance with the Court's settled case-law, to take into account the fact that the person concerned is currently deprived of his liberty and that the question as to whether he may continue to be held in custody depends on the outcome of the dispute in the main proceedings. In addition, the situation of the person concerned must be assessed as it stands at the time when consideration is given to the request that the reference be dealt with under the urgent procedure (judgment of 10 August 2017, Zdziaszek, C271/17 PPU, EU:C:2017:629, paragraph 72 and the case-law cited). |
| 3 | **CELEX: 62018CJ0161**<br>NUMBER: 46<br>TITLE: Violeta Villar Láiz v Instituto Nacional de la Seguridad Social (INSS) and Tesorería General de la Seguridad Social (TGSS).<br>DATE: 2019-05-08<br>TEXT: 46 Further, as is also apparent from recital 30 of Directive 2006/54, the appreciation of the facts from which it may be presumed that there has been indirect discrimination is the task of the national judicial authority, in accordance with national law or national practices which may provide, in particular, that indirect discrimination may be established by any means, and not only on the basis of statistical evidence (see, by analogy, judgment of 19 April 2012, Meister, C415/10, EU:C:2012:217, paragraph 43). | **CELEX: 62010CJ0415**<br>NUMBER: 43<br>TITLE: Galina Meister v Speech Design Carrier Systems GmbH.<br>DATE: 2012-04-19<br>TEXT: 43. In that regard, it should be recalled that, as is clear from Recital 15 of Directives 2000/43 and 2000/78 and Recital 30 of Directive 2006/54, national law or the national practices of the Member States may provide, in particular, that indirect discrimination may be established by any means including on the basis of statistical evidence. |

**Table 11: Examples for scenario (iii): BM25 and SBERT-ft with perfect Recall@20. The first column indicates the example number. We highlight in pink the lexical overlap between the citing and cited paragraphs.**

| | Query document | Cited paragraph |
|---|---|---|
| 1 | **CELEX: 62018CJ0066**<br>NUMBER: 178<br>TITLE: European Commission v Hungary.<br>DATE: 2020-10-06<br>TEXT: 178 As the Court has consistently held, a restriction on the freedom of establishment is permissible only if, in the first place, it is justified by an overriding reason in the public interest and, in the second place, it observes the principle of proportionality, which means that it is suitable for securing, in a consistent and systematic manner, the attainment of the objective pursued and does not go beyond what is necessary in order to attain it (judgment of 23 February 2016, Commission v Hungary, C179/14, EU:C:2016:108, paragraph 166). | **CELEX: 62014CJ0179**<br>NUMBER: 166<br>TITLE: European Commission v Hungary.<br>DATE: 2016-02-23<br>TEXT: 166 The Court has consistently held that such restrictions cannot be justified unless they serve overriding reasons relating to the public interest, are suitable for securing the attainment of the public interest objective which they pursue and do not go beyond what is necessary in order to attain it (see, inter alia, judgments in Läärä and Others, C124/97, EU:C:1999:435, paragraph 31, and OSA, C351/12, EU:C:2014:110, paragraph 70). |
| 2 | **CELEX: 62016CJ0621**<br>NUMBER: 97<br>TITLE: European Commission v Italian Republic.<br>DATE: 2019-03-26<br>TEXT: 97 Moreover, inasmuch as, by that argument, the Commission intends to call into question the General Court's analysis, in paragraphs 110 to 117 of the judgment under appeal, of the content of the general rules governing open competitions, including the general guidelines on the use of languages, and of the notices of competition at issue and the Commission's pleadings before it, it should be recalled that, according to settled case-law, it is apparent from Article 256 TFEU and the first paragraph of Article 58 of the Statute of the Court of Justice of the European Union that the appeal is limited to points of law. The General Court thus has exclusive jurisdiction to find and appraise the relevant facts. The appraisal of those facts thus does not, save where they are distorted, constitute a point of law which is subject, as such, to review by the Court of Justice on appeal (judgment of 8 November 2016, BSH v EUIPO, C43/15 P, EU:C:2016:837, paragraph 50). The Commission did not plead a distortion. | **CELEX: 62015CJ0043**<br>NUMBER: 50<br>TITLE: BSH Bosch und Siemens Hausgeräte GmbH v European Union Intellectual Property Office.<br>DATE: 2016-11-08<br>TEXT: 50 It is settled case-law that, under Article 256 TFEU and the first paragraph of Article 58 of the Statute of the Court of Justice of the European Union, an appeal lies on points of law only. The General Court thus has exclusive jurisdiction to find and appraise the relevant facts. The appraisal of those facts thus does not, save where they are distorted, constitute a point of law which is subject, as such, to review by the Court of Justice on appeal (see, inter alia, judgment of 17 March 2016, Naazneen Investments v OHIM, C252/15 P, not published, EU:C:2016:178, paragraph 59 and the case-law cited). |
| 3 | **CELEX: 62016CJ0586**<br>NUMBER: 89<br>TITLE: Sun Pharmaceutical Industries Ltd, anciennement Ranbaxy Laboratories Ltd and Ranbaxy (UK) Ltd v European Commission.<br>DATE: 2021-03-25<br>TEXT: Sixth, Sun Pharmaceutical erroneously complains that the General Court relied exclusively or principally on the intention of the parties to the agreement at issue in characterising that agreement as a 'restriction by object'. Not only, as the General Court correctly pointed out in paragraphs 212 and 265 of the judgment under appeal, may the characterisation of an agreement as a 'restriction by object' take account of the parties' intention (judgment of 11 September 2014, CB v Commission, C67/13 P, EU:C:2014:2204, paragraph 54 and the case-law cited), but, in addition, neither the General Court nor the Commission relied exclusively or principally on the intention of the parties to the agreement at issue, taking into account principally objective factors, in particular those referred to in paragraphs 72 and 73 of the present judgment. | **CELEX: 62013CJ0067**<br>NUMBER: 54<br>TITLE: Groupement des cartes bancaires (CB) v European Commission.<br>DATE: 2014-09-11<br>TEXT: 54. In addition, although the parties' intention is not a necessary factor in determining whether an agreement between undertakings is restrictive, there is nothing prohibiting the competition authorities, the national courts or the Courts of the European Union from taking that factor into account (see judgment in Allianz Hungária Biztosító and Others (EU:C:2013:160), paragraph 37 and the case-law cited). |

**Table 12: Examples for scenario (iv): BM25 has 100% recall and SBERT-ft has zero Recall@20. The first column indicates the example number. We highlight in pink the lexical overlap between the citing and cited paragraphs.**

individual sought by a European arrest warrant. The consequence of the execution of such a warrant must not be that that individual suffers inhuman or degrading treatment.

CELEX: 62015CJ0404
NUMBER: 104
TITLE: Pál Aranyosi and Robert Căldăraru v Generalstaatsanwaltschaft Bremen.
DATE: 2016-04-05
TEXT: 104 It follows from all the foregoing that the answer to the questions referred is that Article 1(3), Article 5 and Article 6(1) of the Framework Decision must be interpreted as meaning that where there is objective, reliable, specific and properly updated evidence with respect to detention conditions in the issuing Member State that demonstrates that there are deficiencies, which may be systemic or generalised, or which may affect certain groups of people, or which may affect certain places of detention, the executing judicial authority must determine, specifically and precisely, whether there are substantial grounds to believe that the individual concerned by a European arrest warrant, issued for the purposes of conducting a criminal prosecution or executing a custodial sentence, will be exposed, because of the conditions for his detention in the issuing Member State, to a real risk of inhuman or degrading treatment, within the meaning of Article 4 of the Charter, in the event of his surrender to that Member State. To that end, the executing judicial authority must request that supplementary information be provided by the issuing judicial authority, which, after seeking, if necessary, the assistance of the central authority or one of the central authorities of the issuing Member State, under Article 7 of the Framework Decision, must send that information within the time limit specified in the request. The executing judicial authority must postpone its decision on the surrender of the individual concerned until it obtains the supplementary information that allows it to discount the existence of such a risk. If the existence of that risk cannot be discounted within a reasonable time, the executing judicial authority must decide whether the surrender procedure should be brought to an end.

CELEX: 62018CJ0220
NUMBER: 59
TITLE: ML.
DATE: 2018-07-25
TEXT: 59 Accordingly, where the judicial authority of the executing Member State is in possession of information showing there to be a real risk of inhuman or degrading treatment of individuals detained in the issuing Member State, measured against the standard of protection of fundamental rights guaranteed by EU law and, in particular, by Article 4 of the Charter, that judicial authority is bound to assess the existence of that risk when it is called upon to decide on the surrender to the authorities of the issuing Member State of the individual concerned by a European arrest warrant. The consequence of the execution of such a warrant must not be that that individual suffers inhuman or degrading treatment (judgment of 5 April 2016, Aranyosi and Căldăraru, C-404/15 and C-659/15 PPU, EU:C:2016:198, paragraph 88).

CELEX: 62018CJ0220
NUMBER: 62
TITLE: ML.
DATE: 2018-07-25
TEXT: 62 Thus, in order to ensure observance of Article 4 of the Charter in the particular circumstances of a person who is the subject of a European arrest warrant, the executing judicial authority, when faced with evidence of the existence of such deficiencies that is objective, reliable, specific and properly updated, is then bound to determine, specifically and precisely, whether, in the particular circumstances of the case, there are substantial grounds for believing that, following the surrender of that person to the issuing Member State, he will run a real risk of being subject in that Member State to inhuman or degrading treatment, within the meaning of Article 4, because of the conditions for his detention envisaged in the issuing Member State (judgment of 5 April 2016, Aranyosi and Căldăraru, C-404/15 and C-659/15 PPU, EU:C:2016:198, paragraphs 92 and 94).

CELEX: 62018CJ0220
NUMBER: 87
TITLE: ML.
DATE: 2018-07-25
TEXT: 87 Consequently, in view of the mutual trust that must exist between Member States, on which the European arrest warrant system is based, and taking account, in particular, of the time limits set by Article 17 of the Framework Decision for the adoption of a final decision on the execution of a European arrest warrant by the executing judicial authorities, those authorities are solely required to assess the conditions of detention in the prisons in which, according to the information available to them, it is actually intended that the person concerned will be detained, including on a temporary or transitional basis. The compatibility with the fundamental rights of the conditions of detention in the other prisons in which that person may possibly be held at a later stage is, in accordance with the case-law referred to in paragraph 66 of this judgment, a matter that falls exclusively within the jurisdiction of the courts of the issuing Member State.

## E   Quantitative analysis for BM25 and SBERT-ft

We present the comparison between BM25 and SBERT-ft for the quantitative analysis performed in Section 5.2.

The number of $N$-grams in common is statistically significantly higher when both BM25 and SBERT-ft have perfect Recall@20 versus the scenario where SBERT-ft has 100% Recall@20 and SBERT-ft has zero recall, for all $N = 2, 3, ..., 10$. The results for number of $N$-grams in common are displayed in Figure 5.

The mean edit distance in the scenario of perfect recall is 67.23 (std. 38.36) and 109.09 (std. 51.05) in the one BM25 has better performance, where the comparison is statistically significant. The LCS in the perfect recall scenario is 65.86 (std. 31.53) versus 59.23 (std. 28.95) when BM25 has 100% recall and SBERT-ft has zero recall. The results for the mean edit distance and LCS in Figure 6.

Similarly to Ada-v2, the three metrics analyzed point to the same conclusion that there is a lower presence of verbatim quotes and paraphrases between the query and cited documents in the scenario in which BM25 has 100% Recall@20 and Ada-v2 zero recall than in the scenario where both have perfect recall.

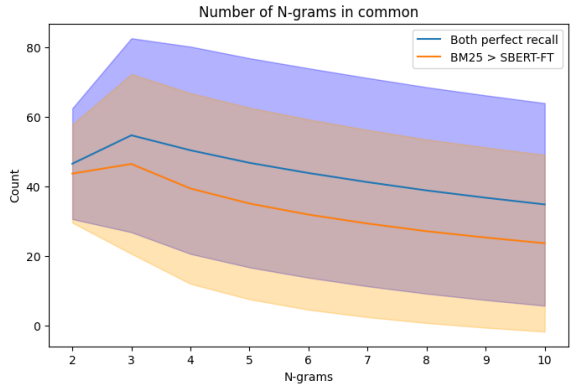

**Figure 5: Average number of $N$-grams in common between query and cited paragraph when both BM25 and SBERT-ft have perfect Recall@20 (*Both perfect recall*), and when BM25 has 100% recall and SBERT-ft has 0% Recall@20 (*BM25 > SBERT-ft*), for $N = 2, 3, ..., 10$. The shaded regions indicate the standard deviation. For all $N$, the average number of $N$-grams in common is statistically significantly higher in the perfect recall scenario.**

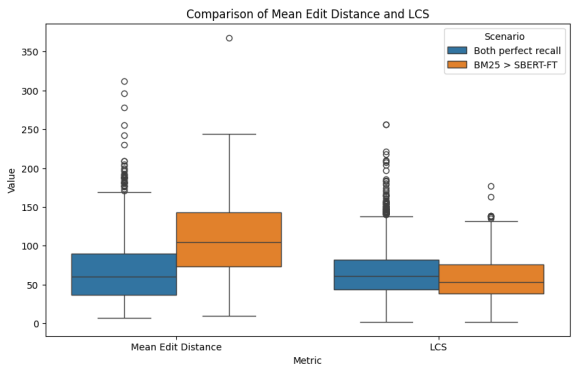

**Figure 6: Comparison between Mean Edit Distance and LCS between query and cited paragraph when both BM25 and SBERT-ft have perfect Recall@20 (*Both perfect recall*), and when BM25 has 100% recall and SBERT-ft has 0% Recall@20 (*BM25 > SBERT-ft*). On average, the mean edit distance is lower and the LCS is higher when both methods have perfect recall, indicating a higher degree of similarity between the documents than when BM25 outperforms Ada-v2. Both results are statistically significant.**

## F Results for different fine-tuning percentages for SBERT-ft

We display the remainder of the evaluation metrics (nDCG@10, MAP and MRR) for different percentages of data used for fine-tuning the SBERT-ft model, with years used for training and validation.

The results show that the models fine-tuned on more data tend to perform better on the test set but performance increases tend to plateau as the amount of data increases. Indeed, for the three metrics shown, the best performing model is the one fine-tuned on 50% of the CJEU data. For instance, the nDCG@10 performance from fine-tuning on 50% of the data is 0.6273, which is higher than the performance of 0.6259 for 70% of data. The same trend is observed is for MAP, where fine-tuning on 50% of data evaluates to 0.5737 versus 0.5724 on 80%.

| Training | Validation | nDCG@10 | MAP | MRR |
|---|---|---|---|---|
| **Zero-shot** | - | 0.4797 | 0.4325 | 0.4921 |
| **10%** 1978-2000 | 2001-2004 | 0.5982 | 0.5450 | 0.6127 |
| **20%** 1978-2004 | 2005-2007 | 0.6152 | 0.5627 | 0.6298 |
| **30%** 1978-2007 | 2008-2010 | 0.6196 | 0.5669 | 0.6340 |
| **50%** 1978-2011 | 2012-2014 | 0.6273 | 0.5737 | 0.6399 |
| **70%** 1978-2015 | 2016-2017 | 0.6259 | 0.5729 | 0.6397 |
| **80%** 1978-2016 | 2017-2018 | 0.6266 | 0.5724 | 0.6391 |
| BM25 | - | 0.5767 | 0.5132 | 0.5786 |

**Table 13: Complementary evaluation metrics (nDCG@10, MAP and MRR) for different percentages of data used for fine-tuning the SBERT-ft model, with years used for training and validation. The test set is the split for 2019-2021 years.**

## G Discussion of cases in which BM25 achieves 0% recall and dense models 100% recall

We analyze the scenario where BM25's performance with zero recall and Ada-v2 has 100% recall and provide more insights into model comparison. In this scenario, the quantitative metrics (e.g., LCS) are closer to the patterns seen when BM25 has 100% recall and Ada-v2 has 0% recall.

However, our qualitative analysis reveals that, when BM25 has 0% recall and Ada-v2 has 100% recall, often the query involves synonyms or a semantic role shift from the cited paragraph. This discrepancy is expected, as contextual models are better suited to capture semantic meaning beyond term overlap. For example, BM25 fails but Ada-v2 succeeds when the query refers to "an individual being entitled to bring an action" whereas the target paragraph discusses the "State being held responsible".

## H Analysis of alternative temporal dataset splits

In Section 3.1, we split the data into training, validation and test sets following a temporal criterion to more closely replicate what would be experienced in legal practice, where practitioners retrieve precedent from past decisions. However, there is the possibility that the period comprising the test split (2019-2021) could introduce a distribution shift that limit the generalizability of our findings to other time periods.

To address this potential issue, we conducted an additional experiment on a broader test set spanning 2015-2021. Specifically, we compared BM25 (statistical model), Ada-vs (zero-shot) and SBERT-ft-50pct (fine-tuned on training data from 1978-2013 and validated on 2012-2014).

| Method | R@1 | R@5 | R@10 | R@20 | nDCG@10 | MAP | MRR |
|---|---|---|---|---|---|---|---|
| BM25 | 0.3743 | 0.6391 | **0.7254** | **0.7843** | 0.5897 | 0.5303 | 0.5944 |
| Ada-vs | 0.3572 | 0.5934 | 0.6726 | 0.7369 | 0.5529 | 0.4997 | 0.5654 |
| SBERT-ft-50pct | **0.4059** | **0.6531** | 0.7244 | 0.7820 | **0.6117** | **0.5591** | **0.6261** |

**Table 14: Performance metrics for a broader test set spanning 2015-2021 than the default 2019-2021 test split. The results largely align with the trends reported in the original manuscript based on the 2019-2021 test split.**

The results largely align with the trends reported in the original manuscript based on the 2019-2021 test split, which are shown in Table 14. The fine-tuned model outperformed the others cross most metrics, except for R@10 and R@20, where BM25 achieved the highest scores. We updated the appendix to the manuscript to include the complete analysis of the additional experiments.